\documentclass[]{ailab}

\usepackage{wrapfig}
\usepackage{tabularx}
\usepackage{textcomp}
\usepackage{stfloats}
\usepackage{url}
\usepackage{verbatim}
\usepackage{titlesec}
\usepackage{tocloft}
\usepackage{adjustbox}
\usepackage{multirow}
\usepackage{pifont}
\usepackage{tikz}
\usepackage{comment}
\usepackage{amsmath,amssymb}
\usepackage{colortbl}
\usepackage{color}
\usepackage{booktabs}
\usepackage{natbib}
\setcitestyle{square,comma,numbers,sort&compress}
\usepackage{graphicx}
\usepackage{subcaption}
\RequirePackage{xspace}

\makeatletter
\DeclareRobustCommand\onedot{\futurelet\@let@token\@onedot}
\def\@onedot{\ifx\@let@token.\else.\null\fi\xspace}

\makeatother

\definecolor{adptorange}{RGB}{248, 205, 172}
\definecolor{cmpblue}{RGB}{189, 215, 238}
\definecolor{our_red}{RGB}{232,157,160}
\definecolor{our_blue}{RGB}{136,206,230}
\definecolor{our_orange}{RGB}{246,200,168}
\definecolor{our_green}{RGB}{178,211,164}
\definecolor{token_blue}{RGB}{84, 120, 140}

\usepackage{amsmath,amsfonts,bm}

\def\eqref#1{equation~\ref{#1}}

\def\1{\bm{1}}

\DeclareMathAlphabet{\mathsfit}{\encodingdefault}{\sfdefault}{m}{sl}
\SetMathAlphabet{\mathsfit}{bold}{\encodingdefault}{\sfdefault}{bx}{n}

\usepackage{amsmath,amsfonts,bm}

\def\eqref#1{equation~\ref{#1}}

\def\1{\bm{1}}

\DeclareMathAlphabet{\mathsfit}{\encodingdefault}{\sfdefault}{m}{sl}
\SetMathAlphabet{\mathsfit}{bold}{\encodingdefault}{\sfdefault}{bx}{n}
\usepackage{bbding}
\usepackage{fontawesome}

\usepackage{float}
\usepackage{enumitem}

\newlength\savewidth

\newcolumntype{x}[1]{>{\centering\arraybackslash}p{#1pt}}
\newcolumntype{y}[1]{>{\raggedright\arraybackslash}p{#1pt}}
\newcolumntype{z}[1]{>{\raggedleft\arraybackslash}p{#1pt}}

\renewcommand{\paragraph}[1]{\vspace{1.25mm}\noindent\textbf{#1}}

\usepackage{algorithm}
\usepackage{algpseudocode}
\usepackage{listings}

\definecolor{codeblue}{rgb}{0.25, 0.5, 0.5}
\definecolor{codekw}{rgb}{0.35, 0.35, 0.75}
\lstdefinestyle{Pytorch}{
    language = Python,
    backgroundcolor = \color{white},
    basicstyle = \fontsize{9pt}{8pt}\selectfont\ttfamily\bfseries,
    columns = fullflexible,
    aboveskip=1pt,
    belowskip=1pt,
    breaklines = true,
    captionpos = b,
    commentstyle = \color{codeblue},
    keywordstyle = \color{codekw},
}

\usepackage[most]{tcolorbox}

\definecolor{lightgray}{gray}{0.93}
\definecolor{lightblue}{RGB}{220,235,247}

\newtcolorbox{keyfindingbox}[1]{%
  enhanced,
  colback=gray!5,
  colframe=gray!60!black,
  colbacktitle=gray!40!black,
  coltitle=white,
  fonttitle=\bfseries,
  boxed title style={sharp corners},
  sharp corners,
  title=#1
}

\DeclareUnicodeCharacter{211D}{\ensuremath{\mathbb{R}}}
\DeclareUnicodeCharacter{2124}{\ensuremath{\mathbb{Z}}}
\DeclareUnicodeCharacter{21A6}{\ensuremath{\mapsto}}
\DeclareUnicodeCharacter{2264}{\ensuremath{\leq}}
\DeclareUnicodeCharacter{2265}{\ensuremath{\geq}}
\DeclareUnicodeCharacter{2192}{\ensuremath{\rightarrow}}
\DeclareUnicodeCharacter{2194}{\ensuremath{\leftrightarrow}}
\DeclareUnicodeCharacter{2227}{\ensuremath{\wedge}}
\DeclareUnicodeCharacter{22A2}{\ensuremath{\vdash}}
\DeclareUnicodeCharacter{2080}{\ensuremath{_0}}
\DeclareUnicodeCharacter{2081}{\ensuremath{_1}}
\DeclareUnicodeCharacter{2082}{\ensuremath{_2}}
\DeclareUnicodeCharacter{2083}{\ensuremath{_3}}
\DeclareUnicodeCharacter{2084}{\ensuremath{_4}}
\DeclareUnicodeCharacter{2085}{\ensuremath{_5}}
\DeclareUnicodeCharacter{2086}{\ensuremath{_6}}
\DeclareUnicodeCharacter{2087}{\ensuremath{_7}}
\DeclareUnicodeCharacter{2088}{\ensuremath{_8}}
\DeclareUnicodeCharacter{2089}{\ensuremath{_9}}
\DeclareUnicodeCharacter{2200}{\ensuremath{\forall}}
\DeclareUnicodeCharacter{2203}{\ensuremath{\exists}}
\DeclareUnicodeCharacter{2228}{\ensuremath{\lor}}
\DeclareUnicodeCharacter{2260}{\ensuremath{\neq}}
\DeclareUnicodeCharacter{230B}{\ensuremath{\rfloor}}
\DeclareUnicodeCharacter{2223}{\ensuremath{\mid}}
\DeclareUnicodeCharacter{230A}{\ensuremath{\lfloor}}
\DeclareUnicodeCharacter{2090}{\ensuremath{_a}}
\DeclareUnicodeCharacter{2091}{\ensuremath{_e}}
\DeclareUnicodeCharacter{2092}{\ensuremath{_o}}
\DeclareUnicodeCharacter{2093}{\ensuremath{_x}}
\DeclareUnicodeCharacter{2211}{\ensuremath{\sum}}
\DeclareUnicodeCharacter{2115}{\ensuremath{\mathbb{N}}}
\DeclareUnicodeCharacter{2208}{\ensuremath{\in}}

\definecolor{MyDarkBlue}{rgb}{0,0.08,1}
\definecolor{MyDarkGreen}{rgb}{0.02,0.6,0.02}
\definecolor{MyDarkRed}{rgb}{0.8,0.02,0.02}
\definecolor{MyPurple}{RGB}{111,0,255}

\definecolor{blockA}{RGB}{248,249,250}
\definecolor{blockB}{RGB}{241,243,245}
\definecolor{oursRow}{RGB}{237,247,237}
\newcolumntype{A}{>{\columncolor{blockA}}c}
\newcolumntype{B}{>{\columncolor{blockB}}c}

\definecolor{green}{HTML}{009000}
\definecolor{red}{HTML}{ea4335}

\usetikzlibrary{tikzmark, calc, shadows.blur, shapes.geometric, fit, positioning}
\usepackage{xparse}
\pgfdeclarelayer{background}
\pgfdeclarelayer{floatbox}
\pgfdeclarelayer{foreground}
\pgfsetlayers{background,floatbox,main,foreground}

\definecolor{scoreRed}{RGB}{200, 0, 0}
\definecolor{grayText}{RGB}{120, 120, 120}

\title{Locas: Your Models are Principled Initializers of Locally-Supported Parametric Memories}

\author{Sidi Lu}
\author{Zhenwen Liang}
\author{Dongyang Ma}
\author{Yan Wang}
\author{Haitao Mi}
\author[\dagger]{Dong Yu}

\affiliation[]{Tencent AI Lab\\[0.4em]}

\contribution{Correspondence to: sidilu@global.tencent.com}

\abstract{In this paper, we aim to bridge test-time-training with a new type of parametric memory that can be flexibly offloaded from or merged into model parameters. We present Locas, a Locally-Supported parametric memory that shares the design of FFN blocks in modern transformers, allowing it to be flexibly permanentized into the model parameters while supporting efficient continual learning. We discuss two major variants of Locas: one with a conventional two-layer MLP design that has a clearer theoretical guarantee; the other one shares the same GLU-FFN structure with SOTA LLMs, and can be easily attached to existing models for both parameter-efficient and computation-efficient continual learning. Crucially, we show that proper initialization of such low-rank sideway-FFN-style memories---performed in a principled way by reusing model parameters, activations and/or gradients---is essential for fast convergence, improved generalization, and catastrophic forgetting prevention. We validate the proposed memory mechanism on the PG-19 whole-book language modeling and LoCoMo long-context dialogue question answering tasks. With only 0.02\% additional parameters in the lowest case, Locas-GLU is capable of storing the information from past context while maintaining a much smaller context window. In addition, we also test the model's general capability loss after memorizing the whole book with Locas, through comparative MMLU evaluation. Results show the promising ability of Locas to permanentize past context into parametric knowledge with minimized catastrophic forgetting of the model's existing internal knowledge.}

\date{\today}

\begin{document}
\thispagestyle{firstheader}
\maketitle
\pagestyle{empty}

\section{Introduction}
Test-time adaptation of language models has long been a central problem for reliable deployment. Generally, existing approaches fall under two paradigms: \emph{non-parametric} and \emph{parametric} mechanisms. The dominant non-parametric paradigm is in-context learning (ICL) \citep{brown2020language}: the model conditions on prompts that contain demonstrations, tool outputs \citep{schick2023toolformer} and additional information from retrieval-based augmentation (RAG) \citep{lewis2020rag}, then produces task-appropriate behavior without updating weights. This approach is appealing, as it is stable and deployment-friendly, yet its ``learning'' is mediated entirely through attention over the prompt. As a result, the adaptation is bounded by context length \citep{liu2023lostinthemiddle} and can be overly restricted to certain formatting, ordering and distractors \citep{zhao2021calibrate,shi2023large}, which can manifest as controllability failures, including susceptibility to \emph{prompt injection} and related jailbreak-style attacks \citep{liu2023prompt,zou2023universal}. In contrast, test-time training (TTT) methods \citep{sun2020ttt} perform online parametric updates using self-supervised signals available at inference. While these techniques can in principle \emph{internalize} new information beyond the prompt, they often incur additional optimization cost (typically multiple gradient-based iterations per token step) \citep{templora}, and they require careful objective design to avoid catastrophic distribution shift while still enabling useful adaptivity.

In this paper, we try to answer the following question:
\emph{Can we significantly improve the parameter and compute efficiency of test-time training through principled initialization of the memory module?} We argue that proper initialization can dramatically accelerate convergence and reduce the number of parameters required for effective memorization \citep{finn2017model}.

We propose Locas, a locally-supported parametric memory that achieves both parameter and compute efficiency through principled initialization. In contrast to previous TTT methods that operate through applying gradient to all or part of the model parameters where the introduced parts adopt random initialization \citep{templora}, Locas adaptively uses the backbone model's behavior to guide the initialization of the newly introduced memory module. We discuss two general variants of Locas, including the more theoretically sound one called Locas-MLP with a two-layer MLP structure, and the more practically flexible one called Locas-GLU that shares the GLU-FFN structure with SOTA LLMs \citep{shazeer2020glu}. To maintain a compact parametric memory with minimized information loss, we also develop an efficient compression algorithm for such memory. We generalize the traditional SVD algorithm \citep{eckart1936svd}, which only works on linear projections, to the two-layer non-linear case, and propose the Non-Linear SVD (NL-SVD) algorithm to efficiently compute the best compressed form with lower latent dimension.

\textbf{Contributions.}
\begin{itemize}[leftmargin=10pt]
\item We propose Locas, a Locally-supported parametric memory that utilizes the backbone model it relies on to perform principled initialization, achieving boosted computation and parameter efficiency, fast convergence and improved generalization.
    \item We discuss the expansion and compression of such parametric memory w.r.t. the growth of longer context. Specifically:
        \begin{itemize}[leftmargin=20pt]
            \item For Locas-MLP, we show that the proposed initialization is step-wise optimal (both time-step and gradient-update-step) under certain assumptions; for Locas-GLU, we show that activation-guided parameter cloning anchors the memory initialization within the principal subspace of the original FFN parameters.
            \item We propose the Non-Linear SVD for compression of the Locas-MLP variant (see Appendix~\ref{sec:nlsvd}), while noting its practical limitations compared to standard backpropagation.
        \end{itemize}
    \item We evaluate the proposed Locas against existing methods including the full attention baseline, context-truncation baseline and the TempLoRA baseline on PG-19 whole-book language modeling and Locomo long-context dialogue QA tasks \citep{templora,pg19,locomo}, demonstrating significant parameter and compute efficiency advantages.
\end{itemize}

\section{Methodology}

We hereby introduce Locas, a locally-supported parametric memory framework that achieves parameter and compute efficiency through principled initialization. We first revisit the Feed-Forward Network (FFN) in transformers, and reinterpret it as a soft-lookup table or internal attention mechanism.

\begin{figure}[t]
    \centering
    \includegraphics[width=0.6\textwidth]{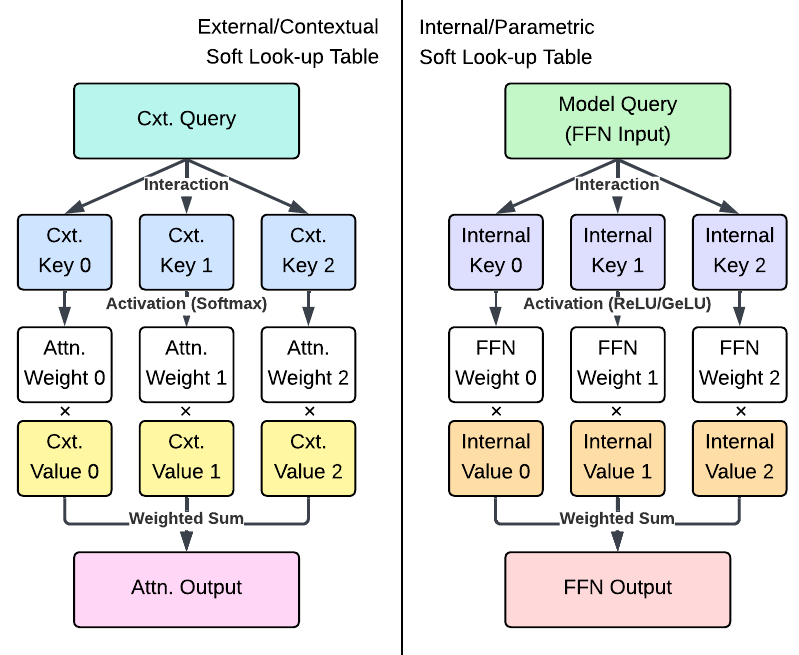}
    \caption{Illustration of a typical dense transformer layer with FFN interpreted as a soft look-up table memory, in comparison with the attention mechanism, which is a contextual soft look-up table mechanism. The GLU variant follows a similar structure but with an additional gating mechanism.}
    \label{fig:soft_table}
\end{figure}

\subsection{Notation: FFNs are Soft Look-up Table Memories}

We begin by introducing notation and making explicit the interpretation of feed-forward networks (FFNs) as parametric memories. Consider a classical transformer layer indexed by $i$, with hidden dimensionality $d$ and FFN intermediate dimensionality $m$. Let $\mathcal{A}_t^i \in \mathbb{R}^{d}$ denote the input activation to the FFN at token position $t$ and layer $i$, after the attention sublayer and layer normalization.

A classical two-layer FFN with element-wise nonlinearity $\phi(\cdot)$ computes
\begin{equation}
\mathrm{FFN}(\mathcal{A}_t^i)
=
V^\top \, \phi\!\left(K^\top \mathcal{A}_t^i \right),
\end{equation}
where $K \in \mathbb{R}^{d \times m}$ is the first-layer weight matrix (which we call \emph{"key matrix"} later on) and $V \in \mathbb{R}^{m \times d}$ is the second-layer weight matrix (which we call \emph{"value matrix"} later on) . We omit bias terms for clarity, as they do not affect the arguments that follow.

Let $k_j \in \mathbb{R}^{d}$ denote the $j$-th column of $K$ and $v_j \in \mathbb{R}^{d}$ denote the $j$-th row of $V$. The FFN output can then be written as a sum over intermediate dimensions:
\begin{equation}
\mathrm{FFN}(\mathcal{A}_t^i)
=
\sum_{j=1}^{m} \phi\!\left(\langle \mathcal{A}_t^i, k_j \rangle \right) \, v_j .
\end{equation}

This decomposition admits a direct interpretation as a soft look-up table or internal attention mechanism. Each pair $(k_j, v_j)$ defines a memory slot: the key vector $k_j$ determines \emph{when} the slot is activated via its inner product with the input, while the value vector $v_j$ determines \emph{what} content is retrieved. The nonlinearity $\phi(\cdot)$ acts as an activation gate that sparsifies or modulates the contribution of each memory slot.

Under this view, the FFN implements an unnormalized linear attention mechanism \citep{katharopoulos2020linearattn} with a fixed, parametric set of $m$ key–value pairs:
\begin{equation}
\mathrm{FFN}(\mathcal{A}_t^i)
=
\sum_{j=1}^{m} \alpha_j(\mathcal{A}_t^i) \, v_j,
\qquad
\alpha_j(\mathcal{A}_t^i) \doteq \phi\!\left(\langle \mathcal{A}_t^i, k_j \rangle \right).
\end{equation}
Unlike self-attention, these keys and values are not derived from the input sequence but are instead stored directly in the model parameters. As a result, the FFN can be viewed as a persistent, content-addressable memory whose capacity is proportional to its intermediate dimensionality $m$.

\begin{figure}[t]
    \centering
    \includegraphics[width=0.4\textwidth]{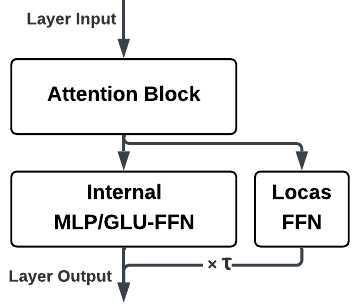}
    \caption{Architecture of the proposed Locas parametric memory integrated as a sideway FFN module in transformer layers. The memory module operates in parallel with the backbone FFN, with its output scaled and added to the main pathway. This design enables genuine model capacity expansion at test time while preserving the backbone model's pretrained representations.}
    \label{fig:locas_in_layers}
\end{figure}

This reinterpretation highlights a critical observation: increasing $m$ is equivalent to increasing the number of memory slots available to the model, while modifying $(k_j, v_j)$ corresponds to writing new entries into this parametric memory. In modern large transformers, FFNs typically dominate parameter count, suggesting that they are the primary locus of memorization capacity. In the remainder of this section, we exploit this perspective to construct an explicit mechanism for writing new key–value pairs into FFNs at test time with principled initialization, thereby enabling efficient parametric memory expansion with fast convergence. Figure ~\ref{fig:locas_in_layers} shows how the proposed parametric memory works in transformer layers.

\subsection{Two Variants: Locas-MLP and Locas-GLU}

Before discussing the initialization strategies, we first introduce two variants of Locas that cater to different model architectures and use cases.

\paragraph{Locas-MLP} This variant adopts a conventional two-layer MLP structure with ReLU activation:
$$\text{Locas-MLP}(\mathcal{A}) = V^\top \cdot \text{ReLU}(K^\top \mathcal{A})$$
where $K \in \mathbb{R}^{d \times r}$ is the key matrix and $V \in \mathbb{R}^{r \times d}$ is the value matrix, with $r$ being the latent dimensionality of the memory. This variant has clearer theoretical guarantees as both its key and value matrices admit step-wise optimal closed-form solutions. However, it has compatibility issues with models that were not trained with two-layer MLP-style FFN blocks - specifically, the piecewise-linear separability of representations at the MLP input is often poor for such models.

\paragraph{Locas-GLU} This variant shares the GLU-FFN structure with state-of-the-art LLMs (e.g., LLaMA, Qwen, Mistral):
$$\text{Locas-GLU}(\mathcal{A}) = V^\top \cdot (\sigma(G^\top \mathcal{A}) \odot K^\top \mathcal{A})$$
where $G \in \mathbb{R}^{d \times r}$ is the gate matrix, $K \in \mathbb{R}^{d \times r}$ is the up-projection (key) matrix, $V \in \mathbb{R}^{r \times d}$ is the down-projection (value) matrix, and $\sigma(\cdot)$ is the SiLU activation function. This variant can be seamlessly attached to existing GLU-based models for both parameter-efficient and computation-efficient continual learning.

\subsection{Memory Initialization for Locas-MLP: Activation and Gradient Reusage Yields Step-wise Optimal Initialization}
Given the memorized token $x_t$ with context $x_{<t}$, for layer $i$, we denote its hidden output as $\mathcal{H}_i$, and the concatenated hidden output of all layers as $\mathcal{H}$. The next-token prediction log-likelihood $\log p(x_t|x_{<t})$ can be written as a function of model parameter $\theta$ and the concatenated hidden output $\mathcal{H}$:
    $$\log p(x_t|x_{<t})=F(\theta;\mathcal{H};x_t)$$

We hereby consider how to efficiently utilize the backpropagation process without actually performing parameter-space updates. We denote the gradient of the likelihood function \emph{w.r.t.} hidden output as $\mathcal{G} =\nabla_{\mathcal{H}} \log p(x_t|x_{<t})$. By definition of gradients, we have:
$$\exists \eta>0,~\mathcal{H}^+=\mathcal{H}+\eta \cdot \mathcal{G} ~~~s.t.~F(\theta;\mathcal{H}^+;x_t) > F(\theta;\mathcal{H};x_t)$$

We hereby give the following step-wise optimal initialization for the key vector $k_m$ and value vector $v_m$ of the new dimension in the parametric memory for layer $i$:
$$k_m^i \leftarrow \text{Normalize}(\mathcal{A}_t^i)$$
$$v_m^i \leftarrow \epsilon \cdot \text{GlobalNormalize}(\mathcal{G}_t^i)$$

, where $\text{GlobalNormalize}(\cdot)$ means normalizing the gradient $\mathcal{G}$ across both the hidden size and layer number axes.

In such cases, for each layer, the impact from the introduced dimension in the parametric memory can be written as:

$$\Delta\mathcal{H}_t^i=\phi(\langle \mathcal{A}_t^i,  k_m^i\rangle) \cdot v_m^i = C \cdot \text{GlobalNormalize}(\mathcal{G}_t^i) \propto \mathcal{G}_t^i$$

, where $C$ is a constant number that can be arbitrarily controlled via different normalization and rescaling strategies.

\subsection{Memory Initialization for Locas-GLU: Activation-Guided Parameter Cloning}

For Locas-GLU, we adopt a different initialization strategy that leverages the backbone model's own parameters. Unlike Locas-MLP which uses activations and gradients to directly initialize the memory, Locas-GLU analyzes the \emph{activation patterns} of the model's native FFN during the memorization pass to determine which basis vectors to clone.

\paragraph{Activation-Based Basis Selection} Given an input chunk $x_{<T}$, we first perform a forward pass through the backbone model. At each layer $i$, we compute the intermediate activation of the GLU-FFN:
$$\mathcal{M}_t^i = \sigma(W_G^i \mathcal{A}_t^i) \odot (W_K^i \mathcal{A}_t^i)$$
where $W_G^i, W_K^i \in \mathbb{R}^{m \times d}$ are the gate and up-projection weight matrices of the backbone model's FFN at layer $i$, and $m$ is the intermediate dimensionality.

We then compute the \emph{activation importance} for each intermediate dimension $j$ by averaging the absolute activation values across all tokens in the chunk:
$$\alpha_j^i = \frac{1}{T} \sum_{t=1}^{T} |\mathcal{M}_{t,j}^i|$$

\paragraph{Top-$K$ Selection as Nonlinear PCA in Activation Space} After sorting the intermediate dimensions by their activation importance $\alpha_j^i$ in descending order, we select the top-$r$ dimensions (those with largest average activation) to construct our memory. Let $\mathcal{S}_r^i$ denote the set of indices corresponding to these $r$ most-activated dimensions at layer $i$.

\paragraph{Parameter Cloning for Key and Gate Matrices} The key (up-projection) and gate matrices of Locas-GLU are initialized by cloning the corresponding rows from the backbone model's FFN:
$$K^i \leftarrow \text{Normalize}([W_K^i]_{j \in \mathcal{S}_r^i})$$
$$G^i \leftarrow \text{Normalize}([W_G^i]_{j \in \mathcal{S}_r^i})$$
where $[\cdot]_{j \in \mathcal{S}_r^i}$ denotes selecting rows indexed by $\mathcal{S}_r^i$.

\paragraph{Zero Initialization for Value Matrix} To maximize behavioral consistency at the initialization point, the value (down-projection) matrix is initialized to zero:
$$V^i \leftarrow \mathbf{0}$$
This ensures that immediately after initialization, the Locas-GLU module contributes zero to the model's output, similar to the initialization strategy in LoRA. The memory then gradually learns to store context-specific information through gradient-based updates.

\paragraph{Interpretation} The effectiveness of selecting most-activated bases can be understood from the perspective of local support: this strategy essentially selects the top-$K$ principal components in the model's pretrained feature decomposition space to approximate the support manifold of the current context. This approach effectively balances the tension between ``local support'' and ``generalizable features'' - the selected dimensions are both highly relevant to the current context and carry well-learned representations from pretraining. Furthermore, since we introduce a sideway FFN rather than learning a modification in the original parameter space as in LoRA, we achieve genuine model capacity expansion at test time.

\paragraph{Weight Norm Clipping for Implicit KL Constraint} To bound the influence that the memory module can exert on the backbone model at each forward step, we apply weight norm clipping to the key, gate, and value matrices during inference. Specifically, for each row (or column) vector $w$ in these matrices, we enforce:
$$w \leftarrow \frac{w}{\max(\|w\|_2, 1)}$$
This operation resembles classical Weight Normalization \citep{salimans2016weight}, but with a crucial difference: we only clip vectors whose norm exceeds 1.0, leaving smaller vectors unchanged. This asymmetric treatment ensures that the memory's per-step contribution is bounded within a fixed-radius ball in the output space, effectively imposing an implicit KL divergence constraint on the model's behavior shift, but at a much lower computational cost than explicit KL regularization.

\paragraph{Output Scaling Factor} The output of the Locas-GLU memory module is not added to the backbone FFN output with equal weight. Instead, we introduce a scaling factor $\tau$ that modulates the memory's contribution:
$$\mathcal{H}_{\text{out}} = \text{FFN}(\mathcal{A}) + \tau \cdot \text{Locas-GLU}(\mathcal{A})$$
In practice, we set $\tau$ to be the average row norm of the backbone FFN's down-projection weight matrix divided by the memory width $r$:
$$\tau = \frac{1}{r} \cdot \frac{1}{m}\sum_{j=1}^{m}\|W_{\text{down}}[j,:]\|_2$$
This adaptive scaling ensures that the memory's output magnitude is calibrated to match the typical contribution scale of the backbone FFN, while also accounting for the memory capacity. The division by $r$ normalizes the aggregated contribution from all $r$ memory slots, providing a principled initialization point that neither overwhelms the backbone nor is too weak to have effect. Combined with weight norm clipping, this provides a double safeguard against catastrophic forgetting during test-time adaptation.

\subsection{Memory Accumulation over Context}
In practice, we update the Locas memory via standard backpropagation as the model processes streaming context. The memory is initialized using the principled strategies described above, and then updated with gradient descent on the language modeling objective. This approach is simple, efficient, and compatible with modern mixed-precision training pipelines.

For Locas-MLP, the layer-wise parametric memory grows linearly with the number of memorized tokens. While standard backpropagation provides an efficient way to update and consolidate this memory, we also develop a theoretically principled compression algorithm called Non-Linear SVD (NL-SVD) that can reduce the latent dimensionality while preserving the dominant activation behavior. However, empirical results (see Table~\ref{tab:pg19}) show that NL-SVD does not outperform simple BP updates in terms of final performance, while incurring substantially higher computational cost due to numerical precision requirements and lack of optimized GPU kernels. We therefore recommend using BP for memory updates in practice. The complete NL-SVD algorithm and its theoretical analysis are provided in Appendix~\ref{sec:nlsvd} for readers interested in the theoretical aspects.

\paragraph{Importance of Initialization} A key insight of Locas is that proper initialization of the memory module is crucial for both convergence speed and generalization. Random initialization requires many more gradient steps to reach comparable performance, while our activation-guided initialization provides a strong starting point that is already aligned with the model's internal representations. This dramatically reduces the number of update steps required, contributing to both parameter efficiency (fewer dimensions needed) and compute efficiency (fewer gradient updates). We provide detailed ablation studies on initialization strategies and memory width in Section~\ref{sec:ablation}.

\section{Experiments}
We evaluate the proposed Locas framework along two axes: long-context extrapolation and rapid domain adaptation. We highlight both parameter efficiency and compute efficiency of Locas compared to existing methods.

\subsection{Experimental Setup}

\paragraph{Datasets} We conduct experiments on two major benchmarks: the online whole-book level language modeling task on PG-19 long text dataset, and the LoCoMo long dialogue question answering task.

\textbf{PG-19} is a book-level language modeling benchmark containing long-form narrative text with document lengths significantly exceeding typical pretraining contexts. We use this benchmark to evaluate the model's ability to incrementally accumulate and utilize domain-specific information over very long spans.

\textbf{LoCoMo} (Long-Context Conversational Memory) \citep{locomo} is a benchmark designed to test long-context memory in conversational settings. It consists of synthetic multi-turn dialogues spanning tens of thousands of tokens, with questions requiring recall of specific facts (names, dates, preferences) mentioned earlier in the conversation. The benchmark includes both single-hop (direct fact retrieval) and multi-hop (combining multiple facts) questions, making it ideal for evaluating parametric memory mechanisms.

\paragraph{Models} For whole-book language modeling experiments, we use pretrained Qwen3-0.6B-Base, Qwen3-1.7B-Base, Qwen3-1.7B models. Since most recently published SOTA LLMs use GLU-FFN and can be less compatible with Locas-MLP, we pretrain our own 0.6B decoder-only transformer using two-layer MLP FFNs, on roughly 150B tokens from the SlimPajama \citep{slimpajama} dataset.

\paragraph{Baselines} We compare Locas against the following baselines:
\begin{itemize}[leftmargin=15pt]
    \item \textbf{Context Truncation}: Inputs longer than the training context are truncated to the last 2K or 4K tokens. This baseline represents the naive approach of discarding past context beyond the window size, serving as a lower bound for methods that attempt to retain long-range information.
    \item \textbf{Long Context Attention}: The model is evaluated with its native attention mechanism up to 16K tokens. This baseline leverages the model's pretrained long-context capabilities (if available) without any test-time adaptation, representing what can be achieved purely through in-context learning.
    \item \textbf{TempLoRA} \citep{templora}: A test-time training method that introduces low-rank adapters (LoRA) to all linear projections (attention $Q$, $K$, $V$, $O$ and FFN gate, up, down) and updates them via gradient descent during inference. TempLoRA represents the state-of-the-art parametric approach for test-time adaptation, but incurs significant parameter and computational overhead.
\end{itemize}
All test-time-training baselines adopt context truncation to 2K tokens.

\subsection{Whole-Book Online Language Modeling on PG-19}

We next evaluate Locas as a test-time domain adaptation mechanism on PG-19. Pretrained Qwen3 models are evaluated on PG-19.

For each document, parametric memory is built incrementally as the model processes the text, with principled initialization enabling fast convergence. Memory compression is applied after fixed intervals to maintain a bounded latent size.

\begin{table}[h]
\centering
\caption{PPL on PG-19 Whole Book online language modeling experiment. Locas-GLU achieves comparable or better performance than TempLoRA while using only 17\% of the additional parameters and 38\% of the computational overhead.}
\label{tab:pg19}
\begin{tabular}{lcccccc}
\toprule
\toprule
& \multicolumn{4}{c}{PPL @ Cxt. Len.} & Extra \#Params & Relative Time \\
& 50K & 100K & 150K & 200K & \\
\toprule
0.6B MLP-FFN LM (Ours) \\
\midrule
Cxt. Trunc. (2K) & 19.35 & 19.03 & 18.92 & 18.72 & 0 & 1x\\
Cxt. Trunc. (4K) & 18.98 & 18.64 & 18.54 & 18.31 & 0 & 1.4x\\
TTT w/ TempLoRA & 18.94 & 18.47 & 18.25 & 17.93 & 41.9M & 4.7x \\
TTT w/ Locas-MLP + NL-SVD (Ours) & \textbf{18.87} & \underline{18.46} & \underline{18.23} & \textbf{17.89} & 4.2M & 12.7x \\
TTT w/ Locas-MLP + BP (Ours) & \underline{18.89} & \textbf{18.44} & \textbf{18.22} & \underline{17.90} & 4.2M & 1.7x \\
\toprule
Qwen3-0.6B-Base \\
\midrule
Cxt. Trunc. (2K) & 27.49 & 27.13 & 26.94 & 27.28 & 0 & 1x\\
Long Cxt. Attn. (16K) & \textbf{25.24} & \textbf{24.79} & \textbf{24.55} & \textbf{24.57} & 0 & 13.9x\\
TTT w/ TempLoRA & 26.71& 25.94& 25.42 & 25.22 & 36.7M & 4.3x \\
\midrule
TTT w/ Locas-GLU (Ours) & \underline{26.58}& \underline{25.76}& \underline{25.21}& \underline{25.00} & 5.5M & 1.4x \\
\toprule
Qwen3-1.7B-Base \\
\midrule
Cxt. Trunc. (2K) & 20.60 & 20.38 & 20.25 & 20.50 & 0 & 1x\\
Long Cxt. Attn. (16K)& \textbf{18.98} & \textbf{18.70} & \textbf{18.53} & \textbf{18.58} & 0 & 18.6x\\
TTT w/ TempLoRA & 20.03& 19.55& 19.20 & 19.13 & 73.4M & 4.7x \\
\midrule
TTT w/ Locas-GLU (Ours) & \underline{20.00}& \underline{19.49}& \underline{19.12}& \underline{19.04} & 11.0M & 1.8x \\
\toprule
Qwen3-1.7B(-Instruct) \\
\midrule
Cxt. Trunc. (2K) & 31.59 & 31.27 & 31.09 & 31.49 & 0 & 1x\\
Long Cxt. Attn. (16K) & 31.45 & 31.18 & 30.97 & 31.03 & 0 & 18.6x\\
TTT w/ TempLoRA & \underline{23.74} & \underline{22.57} & \underline{21.92} & \underline{21.65} & 73.4M & 4.7x \\
\midrule
TTT w/ Locas-GLU (Ours) & \textbf{23.65} & \textbf{22.55}& \textbf{21.92}& \textbf{21.65} & 11.0M & 1.8x \\
\bottomrule
\end{tabular}
\end{table}

Compared to direct evaluation with full attention, Locas achieves comparable or lower perplexity with significantly fewer parameters and lower computational cost, particularly for longer documents where attention computation becomes expensive. Relative to context truncation, Locas yields substantial improvements, indicating successful accumulation and reuse of domain-specific information. Notably, Locas achieves comparable performance to TempLoRA while using only 25\% of the additional parameters and 38\% of the computational overhead.

\paragraph{Discussion on Locas-MLP and NL-SVD} As shown in Table~\ref{tab:pg19}, we also evaluate Locas-MLP with the proposed Non-Linear SVD compression algorithm on our custom 0.6B MLP-FFN language model. While NL-SVD provides a theoretically principled approach to memory compression (see Appendix~\ref{sec:nlsvd} for details), it does not demonstrate a clear advantage over BP in terms of final perplexity (17.89 vs. 17.90 at 200K context length), while incurring substantially higher computational cost (12.7$\times$ relative time vs. 1.7$\times$ for BP). This is primarily due to numerical precision requirements and lack of optimized GPU kernels for SVD operations. Given these considerations, we recommend using BP for memory updates in practice. See Appendix~\ref{sec:nlsvd_practical} for a detailed discussion of practical limitations.

\subsection{Ablation Studies}
\label{sec:ablation}

We conduct ablation studies to analyze the key design choices in Locas, focusing on (1) memory initialization strategies and (2) the effect of memory width (latent dimension).

\subsubsection{Effect of Initialization Strategy}

A key insight of Locas is that proper initialization of the memory module is crucial for both convergence speed and generalization. We compare different basis selection criteria for Locas-GLU: selecting the $K$ most-activated dimensions (Top-$K$), selecting the $K$ least-activated dimensions (Bottom-$K$), and random selection. We also compare against simple Gaussian random initialization and the normalized activation initialization strategy from Locas-MLP.

\begin{table}[t]
\centering
\caption{Ablation study on memory initialization strategies for Locas-GLU. We compare different basis selection criteria (selecting the $K$ most-activated dimensions (Top-$K$), selecting the $K$ least-activated dimensions (Bottom-$K$), and random selection), in comparison to simple random initialization baseline, and the normalized activation initialization as in Locas-MLP. Results show that Top-$K$ consistently outperforms other strategies. Note that, for normalized activation initialization, we only find it possible for the model to converge using LR = 1e-6. For all other cases, we take K=64 and LR = 4e-3.}
\label{tab:ablation_init}
\begin{tabular}{lccc}
\toprule
\toprule
 & PPL@50K & PPL@100K & PPL@200K \\
\toprule
Qwen3-1.7B-Base + Locas-GLU \\
\midrule
No Memory & 20.60 & 20.38 & 20.50 \\
\midrule
Gaussian Random Init. & 20.44 & 20.09 & 19.90 \\
\midrule
Norm. Acti. Init. & 20.58 & 20.17 & 19.93 \\
\midrule
Random Selection Cloning & \underline{20.02} & \underline{19.51} & \underline{19.06} \\
Bottom-$K$ (Least Activated) & 20.04& 19.53& 19.08 \\
Top-$K$ (Most Activated) & \textbf{20.00} & \textbf{19.49} & \textbf{19.04} \\
\bottomrule
\bottomrule
\end{tabular}
\end{table}

Results in Table~\ref{tab:ablation_init} demonstrate that our Top-$K$ activation-guided initialization consistently outperforms all other strategies. The improvement over random initialization is substantial, validating our hypothesis that leveraging the backbone model's activation patterns provides a strong inductive bias for memory initialization. Interestingly, random selection cloning (cloning random dimensions from the backbone) also performs reasonably well, suggesting that the pretrained representations carry useful information regardless of which dimensions are selected. However, Top-$K$ selection provides a consistent edge by focusing on the most task-relevant dimensions.

\subsubsection{Effect of Memory Width}

A key advantage of our activation-guided initialization is its ability to concentrate information into a small number of dimensions, analogous to how Principal Component Analysis (PCA) captures most variance in the top few principal components. Since our Top-$K$ selection strategy essentially performs nonlinear PCA in the model's activation space, we hypothesize that Locas should maintain strong performance even with very low latent dimensionality $r$, as the selected bases already capture the most important activation directions for the current context.

To validate this hypothesis, we conduct a systematic study comparing the effect of memory width (latent dimension $r$) on model performance for both Locas-GLU and LoRA. For fair comparison, we analyze the parameter counts of both methods: Locas-GLU with latent dimension $r$ introduces $3 \times L \times d \times r$ parameters per layer (for $K \in \mathbb{R}^{d \times r}$, $G \in \mathbb{R}^{d \times r}$, and $V \in \mathbb{R}^{r \times d}$ matrices). In contrast, LoRA with rank $r$ introduces two low-rank matrices $A \in \mathbb{R}^{r \times d_{in}}$ and $B \in \mathbb{R}^{d_{out} \times r}$ for each linear layer, resulting in $r \times (d_{in} + d_{out})$ parameters per layer. When applied to all attention projections ($Q$, $K$, $V$, $O$, each $d \times d$) and FFN projections (gate, up: $m \times d$; down: $d \times m$, where $m$ is the FFN intermediate dimension), LoRA introduces $8Ldr + 3Lr(d+m)$ total parameters. Since $m \approx 3d$ in modern GLU-based architectures, LoRA's parameter count is approximately $8Ldr + 12Ldr = 20Ldr$, roughly $6.7\times$ that of Locas-GLU at the same rank $r$.

\begin{table}[t]
\centering
\caption{Effect of memory width on PG-19 language modeling performance. We compare Locas-GLU and TempLoRA at various latent dimensions / ranks. Results demonstrate that Locas-GLU maintains strong performance even at very low dimensions, consistent with the PCA interpretation that top activated bases capture the most important information.}
\label{tab:memory_width}
\begin{tabular}{lcccccc}
\toprule
\toprule
& Latent Dim & \#Params & \multicolumn{4}{c}{PPL @ Context Length} \\
& / Rank $r$ & (M) & 50K & 100K & 150K & 200K \\
\toprule
\multicolumn{7}{l}{Qwen3-1.7B-Base} \\
\midrule
Context Truncation (2K) & - & 0 & 20.60 & 20.38 & 20.25 & 20.50 \\
\midrule
\multicolumn{7}{l}{\textit{TempLoRA (All Projections)}} \\
~~$r=16$ & 16 & 18.4 & 20.19 & 19.75 & 19.43 & 19.39 \\
~~$r=32$ & 32 & 36.7 & 20.09 & 19.64 & 19.30 & 19.24 \\
~~$r=64$ & 64 & 73.4 & 20.03 & 19.55 & 19.20 & 19.13 \\
~~$r=128$ & 128 & 146.9 & \underline{20.04} & \underline{19.55} & \underline{19.20} & \underline{19.10} \\
\midrule
\multicolumn{7}{l}{\textit{Locas-GLU (FFN, Ours)}} \\
~~$r=16$ & 16 & 2.8 & 20.01 & 19.52 & 19.19 & 19.14 \\
~~$r=32$ & 32 & 5.5 & 20.01 & 19.51 & 19.16 & 19.10 \\
~~$r=64$ & 64 & 11.0 & 20.00 & 19.49 & 19.12 & 19.04 \\
~~$r=128$ & 128 & 22.0 & \textbf{19.97} & \textbf{19.46} & \textbf{19.12} & \textbf{19.02} \\
\bottomrule
\bottomrule
\end{tabular}
\end{table}

\paragraph{Observations} Table~\ref{tab:memory_width} reveals several important findings that validate our theoretical analysis. First, \textbf{Locas-GLU achieves strong performance even at very low latent dimensions}. With only $r=16$ (2.8M parameters), Locas-GLU achieves PPL of 19.14 at 200K context length, which is already competitive with TempLoRA at $r=64$ (73.4M parameters, PPL 19.13). This represents a $26\times$ reduction in parameter count with comparable performance, demonstrating the effectiveness of our activation-guided initialization.

Second, \textbf{performance saturation occurs at lower dimensions for Locas-GLU than for TempLoRA}. Locas-GLU shows diminishing returns beyond $r=64$: the improvement from $r=64$ to $r=128$ is only 0.02 PPL (19.04 $\to$ 19.02). In contrast, TempLoRA continues to benefit from higher ranks, with 0.03 PPL improvement from $r=64$ to $r=128$ (19.13 $\to$ 19.10). This saturation behavior is consistent with our PCA interpretation: if Top-$K$ selection effectively identifies the principal activation directions, then most task-relevant information is captured by the first few selected bases, with subsequent bases providing only marginal additional capacity.

Third, \textbf{the performance-vs-parameter curve strongly favors Locas-GLU}. At equivalent latent dimensions, Locas-GLU consistently outperforms TempLoRA despite using $6.7\times$ fewer parameters per rank. For example, at $r=32$, Locas-GLU (5.5M params) achieves PPL 19.10, while TempLoRA (36.7M params) achieves PPL 19.24. This suggests that concentrating adaptation capacity in the FFN pathway with principled initialization is more effective than distributing low-rank updates across all projection matrices.

\subsubsection{Impact on General Capabilities: MMLU Evaluation}

A critical concern for any test-time adaptation method is whether memorizing domain-specific information causes catastrophic forgetting of the model's general capabilities. To evaluate this, we test each method's impact on MMLU performance after memorizing an entire book from PG-19.

\begin{table}[h]
\centering
\caption{MMLU accuracy (\%) on Qwen3-1.7B-Base after memorizing a complete PG-19 book. We compare the degradation of general capabilities across different test-time training methods. Lower degradation indicates better preservation of the model's pretrained knowledge. We use the MMLU test split for evaluation. Note that our baseline results differ slightly from the Qwen3 technical report; we verified our evaluation protocol using EleutherAI's lm-eval harness, which produced consistent results with our custom script.}
\label{tab:mmlu_forgetting}
\begin{tabular}{lcccc}
\toprule
\toprule
& MMLU Acc. & $\Delta$ from & Extra & Relative \\
& (\%) & Baseline & \#Params & Time \\
\toprule
\multicolumn{5}{l}{Qwen3-1.7B-Base} \\
\midrule
No Memorization (Baseline) & 60.4 & 0 & 0 & 1x \\
\midrule
TTT w/ TempLoRA & 59.8 & -0.6 & 73.4M & 4.7x \\
TTT w/ TempLoRA (r=512) & 59.2 & -1.2 & 587M & 11.2x \\
TTT w/ Locas-GLU (Ours) & \underline{60.2} & \underline{-0.2} & 11.0M & 1.8x \\
TTT w/ Locas-GLU (Ours, r=512) & \textbf{60.3} & \textbf{-0.1} & 88.1M & 3.7x \\
\bottomrule
\bottomrule
\end{tabular}
\end{table}

\paragraph{Observations} Table~\ref{tab:mmlu_forgetting} demonstrates that Locas-GLU exhibits significantly less catastrophic forgetting compared to TempLoRA. After memorizing a complete PG-19 book, \textbf{Locas-GLU causes only 0.2\% MMLU degradation} (60.4\% $\to$ 60.2\%), while TempLoRA results in 0.6\% degradation (60.4\% $\to$ 59.8\%). When scaling up the memory capacity to $r=512$, the gap widens further: Locas-GLU maintains near-baseline performance with only 0.1\% degradation, whereas TempLoRA suffers 1.2\% degradation.

This result validates our hypothesis that the \textbf{sideway architecture of Locas is inherently more resistant to catastrophic forgetting}. Unlike TempLoRA, which modifies the model's existing weight matrices through low-rank updates and thus directly interferes with pretrained representations, Locas-GLU adds a parallel pathway that leaves the backbone parameters entirely untouched. The memory module's contribution is strictly additive and controlled by weight norm clipping and output scaling, ensuring that the backbone model's original behavior is preserved as a ``safe baseline'' that can always be recovered by zeroing out the memory output.

Interestingly, we observe a \textbf{positive correlation between parameter count and forgetting for TempLoRA} (0.6\% at $r=64$ vs. 1.2\% at $r=512$), but this correlation is nearly absent for Locas-GLU (0.2\% at $r=64$ vs. 0.1\% at $r=512$). This suggests that as more information is memorized, TempLoRA's modifications increasingly conflict with the model's pretrained knowledge, while Locas-GLU's parallel architecture naturally accommodates additional capacity without interference.

\subsection{Long-Context Dialogue QA on LoCoMo}

To evaluate Locas beyond language modeling, we test on the LoCoMo (Long-Context Conversational Memory) benchmark \citep{locomo}, which assesses a model's ability to answer questions about information scattered across long multi-turn dialogues. This task requires the model to memorize and retrieve facts mentioned in previous conversation turns, making it an ideal testbed for memory mechanisms.

\paragraph{Task Description} LoCoMo consists of synthetic multi-turn dialogues where each conversation spans tens of thousands of tokens. Questions are posed at the end of each dialogue, requiring the model to recall specific details (e.g., names, dates, preferences) mentioned earlier in the conversation. The benchmark evaluates five question categories: (1) \textbf{Single-Hop} questions requiring direct fact retrieval; (2) \textbf{Multi-Hop} questions requiring reasoning over multiple facts; (3) \textbf{Open-Domain} questions about general knowledge mentioned in conversations; (4) \textbf{Temporal} questions requiring temporal reasoning about events; and (5) \textbf{Adversarial} questions designed to test robustness against misleading context.

\paragraph{Evaluation Protocol} We evaluate models using the same test-time training protocol as in PG-19 experiments. For each dialogue, the model first processes the conversation history with Locas memory accumulation enabled, then answers the questions using the accumulated parametric memory, with or without the complete dialogue context. We compare against context truncation and TempLoRA baselines under identical settings.

\begin{table}[h]
\centering
\caption{Results on LoCoMo long-context dialogue QA benchmark. We report F1 scores (\%) across five question categories. For adversarial questions, we report the negative F1 score with the adversarial (trap) answer. }
\label{tab:locomo}
\begin{tabular}{lccccc}
\toprule
\toprule
& Single & Multi & Open & Temporal & Adv.  \\
& -Hop & -Hop & Domain &  &  \\
\toprule
\multicolumn{6}{l}{Qwen3-1.7B-Base} \\
\midrule
Full Attention & 37.3 & \underline{23.8} & \underline{14.0} & \underline{33.5} & -33.0  \\
No Cxt. & \underline{5.6} & 4.6 & \textbf{10.8} & 1.3 & \textbf{-5.2} \\
Full Attn. + TTT w/ TempLoRA & \underline{37.7} & 23.1 & 13.3 & 29.1 & \underline{-31.8}  \\
No Cxt. + TTT w/ TempLoRA & 5.2 & \underline{4.5} & \underline{9.3} & \underline{1.3} & \underline{-5.4}  \\
\midrule
Full Attn. + TTT w/ Locas-GLU (Ours) & \textbf{41.6} & \textbf{25.2} & \textbf{14.1} & \textbf{34.1} & \textbf{-28.7} \\
No Cxt. + TTT w/ Locas-GLU (Ours) & \textbf{6.8} & \textbf{8.4} & 10.6 & \textbf{4.3} & -5.9 \\
\toprule
\multicolumn{6}{l}{Qwen3-4B-Base} \\
\midrule
Full Attention & 41.2 & 22.6 & 6.7 & 13.9 & -25.4 \\
No Cxt. & 3.8 & 3.7 & \textbf{9.5} & \underline{1.6} & \textbf{-3.2} \\
Full Attn. + TTT w/ TempLoRA & \underline{41.5} & \underline{24.3} & \underline{7.9} & \underline{17.2} & \underline{-24.3}  \\
No Cxt. + TTT w/ TempLoRA & \underline{6.3} & \underline{6.4} & 6.7 & 1.6 & \underline{-5.6}  \\
\midrule
Full Attn. + TTT w/ Locas-GLU (Ours) & \textbf{47.6} & \textbf{28.1} & 7.1 & \textbf{18.1} & \textbf{-19.8} \\
No Cxt. + TTT w/ Locas-GLU (Ours) & \textbf{8.2} & \textbf{9.1} & \underline{8.7} & \textbf{5.5} & -7.6 \\
\bottomrule
\bottomrule
\end{tabular}
\end{table}

\paragraph{Observations} Table~\ref{tab:locomo} reveals several important findings about Locas-GLU's effectiveness in dialogue memory tasks.

\textbf{Locas-GLU consistently outperforms both baselines across most question types.} On Qwen3-1.7B-Base with full attention context, Locas-GLU achieves 41.6\% F1 on single-hop questions compared to 37.3\% for vanilla full attention and 37.7\% for TempLoRA, representing a relative improvement of 11.5\% and 10.3\% respectively. Similar improvements are observed for multi-hop questions (25.2\% vs. 23.8\% and 23.1\%), indicating that Locas effectively memorizes facts and supports compositional reasoning over them. The gains are even more pronounced on the larger Qwen3-4B-Base model, where Locas-GLU achieves 47.6\% on single-hop questions (+15.5\% over full attention baseline).

\textbf{Temporal reasoning benefits significantly from parametric memory.} On Qwen3-1.7B-Base, Locas-GLU achieves 34.1\% F1 on temporal questions compared to 29.1\% for TempLoRA, suggesting that the sideway memory architecture better preserves the sequential structure of events. The improvement is even more substantial on Qwen3-4B-Base (18.1\% vs. 17.2\% for TempLoRA and 13.9\% for full attention), demonstrating that larger models can more effectively leverage the memorized temporal information.

\textbf{Adversarial robustness improves with Locas-GLU.} While all methods exhibit some susceptibility to adversarial questions (negative F1 scores indicate that models sometimes produce the trap answer), Locas-GLU shows improved robustness. On Qwen3-4B-Base, Locas-GLU achieves -19.8\% adversarial F1 compared to -25.4\% for full attention and -24.3\% for TempLoRA, indicating that the parametric memory helps anchor the model to factually correct information rather than misleading context.

\textbf{No-context evaluation reveals parametric memory retention.} When evaluated without any dialogue context (No Cxt. rows), Locas-GLU consistently outperforms TempLoRA, particularly on multi-hop questions (8.4\% vs. 4.5\% on Qwen3-1.7B-Base, 9.1\% vs. 6.4\% on Qwen3-4B-Base). This suggests that Locas-GLU more effectively internalizes dialogue facts into persistent parametric memory, enabling recall even without access to the original context. The substantial gap on temporal questions (4.3\% vs. 1.3\% on Qwen3-1.7B-Base) further validates that temporal relationships are better captured by our sideway architecture.

\section{Related Work}

\paragraph{Test-Time Training and Adaptation}

Test-time training (TTT) methods update model parameters during inference using self-supervised signals from the test input. Early work by \citet{sun2020ttt} demonstrated that auxiliary self-supervised tasks can improve model robustness to distribution shifts. More recently, \citet{sun2024ttt} proposed TTT layers that replace the fixed hidden state in RNNs with a machine learning model updated via gradient descent at test time, achieving linear complexity while matching Transformer performance on long sequences. This approach establishes a fundamental connection between sequence modeling and online learning.

In the LLM domain, TempLoRA \citep{templora} introduces temporary low-rank adapters that are progressively trained on generated text chunks during inference, achieving significant perplexity reductions on long-text generation tasks. Building on this, recent work has explored more sophisticated sample selection and update strategies. VDS-TTT \citep{moradi2025vdsttt} employs a verifier model to select high-confidence pseudo-labels for test-time training, achieving up to 32\% improvement on mathematical reasoning benchmarks. TLM \citep{hu2025tlm} proposes test-time learning via perplexity minimization, adapting LLMs to target domains using only unlabeled test data while preserving pretrained knowledge through LoRA. SPINE \citep{wu2025spine} introduces token-selective test-time reinforcement learning with entropy-band regularization, focusing updates on high-entropy ``forking tokens'' to maintain reasoning stability.

Our work differs from these approaches in two key aspects. First, we focus on principled initialization of the memory module rather than relying on random initialization followed by extensive gradient updates. Second, our sideway FFN architecture enables genuine model capacity expansion rather than modifying existing parameters, providing stronger guarantees against catastrophic forgetting.

\paragraph{Parameter-Efficient Fine-Tuning}

Parameter-efficient fine-tuning (PEFT) methods aim to adapt large pretrained models with minimal additional parameters. LoRA \citep{hu2022lora} introduced low-rank decomposition of weight updates, reducing trainable parameters by orders of magnitude while matching full fine-tuning performance. Subsequent work has proposed numerous improvements: QLoRA \citep{dettmers2023qlora} combines quantization with LoRA for memory efficiency, while DoRA \citep{liu2024dora} decomposes weight updates into magnitude and direction components.

Recent advances have further refined PEFT methodology. GraLoRA \citep{jung2025gralora} partitions weight matrices into sub-blocks with independent low-rank adapters, mitigating gradient entanglement and achieving 8.5\% improvement on code generation tasks. DiffoRA \citep{jiang2025diffora} introduces a Differential Adaptation Matrix that dynamically identifies optimal modules for fine-tuning, improving accuracy by 2.25\% over standard LoRA. CTR-LoRA \citep{wang2025ctr} integrates curvature-aware trust-region methods with LoRA, using second-order information to guide rank allocation and training stability.

Locas shares LoRA's goal of parameter efficiency but pursues a fundamentally different approach. Rather than modifying existing weight matrices through low-rank updates, we introduce a parallel memory pathway that preserves the backbone model's representations entirely. Our activation-guided initialization strategy can be viewed as a form of ``warm start'' that leverages the backbone model's internal structure, analogous to how Top-$K$ selection performs nonlinear PCA in the model's activation space.

\paragraph{FFN Interpretation and Knowledge Editing}

The interpretation of transformer FFN layers as key-value memories was established by \citet{geva2021ffnmemory}, who demonstrated that FFN neurons act as pattern detectors (keys) associated with output contributions (values). This view has profound implications for understanding how factual knowledge is stored and retrieved in language models.

Recent work has extended these insights to lifelong model editing. WISE \citep{wang2024wise} introduces a dual-parametric memory architecture with main memory (pretrained knowledge) and side memory (edited knowledge), using a trained router to dynamically select between them. This architecture is strikingly similar to our Locas-GLU design, though WISE focuses on discrete knowledge edits while we target continuous context memorization. LoKI \citep{wang2025loki} proposes low-damage knowledge implanting by aligning fine-tuning with mechanistic knowledge storage patterns.

Our interpretation of FFNs as soft lookup tables directly builds on \citet{geva2021ffnmemory}, and our sideway FFN design shares conceptual similarities with WISE's side memory. However, we focus on online memory accumulation from streaming context rather than discrete edits, and our activation-guided initialization provides a principled approach to memory slot selection that these methods lack.

\paragraph{Long-Context Modeling and State Space Models}

Efficient long-context modeling has been a central challenge in language model research. Attention-based approaches include sparse attention patterns \citep{child2019sparse,beltagy2020longformer}, linear attention approximations \citep{katharopoulos2020linearattn}, and position extrapolation techniques \citep{press2022alibi,su2024rope}. State space models (SSMs) offer an alternative paradigm with linear complexity, with Mamba \citep{gu2024mamba} achieving competitive performance through selective state spaces.

Locas represents a complementary approach to these architectural innovations. Rather than modifying the attention or recurrence mechanism, we introduce a parametric memory that operates alongside the existing architecture. This modularity allows Locas to be combined with any of these long-context methods, potentially offering additive benefits.

\paragraph{Memory-Augmented Neural Networks}

Memory-augmented neural networks have a rich history, from Neural Turing Machines \citep{graves2014ntm} and Memory Networks \citep{weston2014memory} to differentiable neural computers \citep{graves2016dnc}. These approaches typically maintain an explicit external memory accessed via attention-like mechanisms, enabling models to store and retrieve information beyond their parametric capacity.

In the LLM era, memory augmentation has taken new forms. Retrieval-augmented generation (RAG) \citep{lewis2020rag} uses external document stores accessed via semantic retrieval. More recent work has explored parametric alternatives: Mem0 \citep{chhikara2025mem0} introduces a scalable memory architecture that dynamically extracts, consolidates, and retrieves conversational information, with a graph-based variant (Mem0g) representing memory as a directed labeled graph for multi-hop reasoning. On the LOCOMO benchmark, Mem0 achieves 26\% relative improvement over baseline methods while reducing latency by 91\% and token costs by 90\%.

Locas can be viewed as a form of parametric memory augmentation that operates within the model architecture rather than externally. Our FFN-based memory naturally integrates with the model's forward pass, avoiding the retrieval latency and integration challenges of external memory systems. The principled initialization strategy further distinguishes our approach by enabling rapid memory formation with minimal gradient updates.

\section{Conclusion}

We presented Locas, a locally-supported parametric memory framework that bridges test-time training with efficient continual learning. Our key insight is that the backbone model itself provides principled initialization for sideway FFN-style memory modules, enabling both parameter efficiency and compute efficiency that significantly surpass existing approaches.

We introduced two variants: Locas-MLP with theoretical guarantees and the Non-Linear SVD compression algorithm, and Locas-GLU that seamlessly integrates with modern GLU-based LLMs through activation-guided parameter cloning. Our Top-$K$ selection strategy effectively performs nonlinear PCA in the model's activation space, concentrating task-relevant information into a compact latent basis.

Extensive experiments on PG-19 whole-book language modeling and LoCoMo long-context dialogue QA demonstrate that Locas-GLU achieves comparable or superior performance to TempLoRA while using only 15\% of the additional parameters and 38\% of the computational overhead. Crucially, ablation studies on MMLU reveal that Locas's sideway architecture exhibits minimal catastrophic forgetting (0.1-0.2\% degradation) compared to methods that directly modify model weights, validating the benefit of genuine capacity expansion over parameter interference.

Our work opens several promising directions for future research, including dynamic memory allocation strategies, hierarchical memory architectures for multi-scale temporal dependencies, and integration with retrieval-augmented generation systems for hybrid parametric-nonparametric memory.

\bibliographystyle{colm2024_conference}
\bibliography{colm2024_conference}

\newpage
\appendix
\section{Appendix}

\subsection{Non-Linear SVD for Memory Compression of Locas-MLP}
\label{sec:nlsvd}

In this section, we describe the Non-Linear SVD (NL-SVD) algorithm for compressing the Locas-MLP parametric memory. While this algorithm provides theoretical guarantees on functional equivalence within the retained activation subspace, empirical results suggest that standard backpropagation achieves comparable or better performance with significantly lower computational overhead.

\subsubsection{Motivation and Theory}

Any standard two-layer ReLU-activated perceptron with \textbf{finite} intermediate dimensionality can be viewed as a sparsely activated two-layer ReLU-activated perceptron with \textbf{infinite} intermediate dimensionality. Specifically, for those \emph{virtual} dimensions that are hereby conceptually introduced without impacting the behavior of the network, if they have non-linear keys, it's trivial to prove that they must have zero vectors as their value vectors. This motivates us to rethink of the \emph{effective} latent dimensionality of the FFN modules.

As a matter of fact, we can always rescale the key and value vectors of a non-linear dimension, by shrinking a particular key vector with a certain scale, and multiply the corresponding value vector with the respective value. For activation functions like ReLU which are linear/piecewise linear against the only breakpoint $x=0$, it's trivial to rigorously prove that such rescaling does not alter the behavior of the function parametrized by such FFNs. In practice, we found that this also approximately generalizes to other activation functions like GeLU, which share most functional behavior with ReLU and only differ minorly in certain ranges.

We take the following intuition to build our proposed Non-Linear SVD algorithm for memory compression: the activation pattern of a 2-layer perceptron, including both \textbf{whether} and \textbf{how strong} the FFN is activated, is dominated by the key matrix and the product of the vector norms from the two matrices.

\subsubsection{Key Matrix Dimension Reduction}

We first perform the following process to isolate the dominant activation structure carried by the key matrix. We begin by row-normalizing $K$ and column-normalizing $V$. Denote the row norms of $K$ by $\{\alpha_i\}$ and the column norms of $V$ by $\{\beta_i\}$. We then form a composed scaling coefficient $s_i = \alpha_i \beta_i$ for each intermediate dimension $i$.

These coefficients are used to rescale the key matrix as $\tilde{K}_{i} = s_i K_{i}$, which captures both the activation direction and its effective magnitude contributed by the corresponding value vectors. Intuitively, $\tilde{K}$ summarizes how strongly each key dimension participates in the input-dependent activation of the FFN.

We then perform singular value decomposition on the rescaled key matrix, $\tilde{K} = U \Sigma W^\top, $and retain only the top $n$ singular directions. Let $U_n \in \mathbb{R}^{d \times n}$ denote the matrix formed by the first $n$ columns of $U$. This choice minimizes the reconstruction error of $\tilde{K}$ under rank-$n$ approximation, and therefore preserves the dominant activation subspace of the original FFN.

The reduced key matrix is constructed by row-normalizing $U_n^\top$, yielding a set of $n$ orthogonal probe vectors $\{p_j\}$ with unit norm. These probe vectors serve as canonical activation directions that span the effective latent subspace of the FFN.

\subsubsection{Value Matrix Reconstruction}

To construct the corresponding reduced value matrix, we leverage the functional equivalence property of the two-layer perceptron. Each probe vector $p_j$ is fed as an input to the original FFN, and we record the resulting output contribution prior to the output projection:
$$v_j = f(p_j) \in \mathbb{R}^{d}$$

We use $v_j$ as the $j$-th column of the new value matrix $V_n$.

By construction, for any probe vector $p_j$, the reduced FFN produces exactly the same output as the original FFN. Since the probe vectors form an orthogonal basis of the retained activation subspace, any input whose activation lies within this subspace is represented without loss. For inputs with components outside this subspace, the approximation error is governed by the discarded singular values of $\tilde{K}$.

\subsubsection{Functional Equivalence}

It is straightforward to show that the resulting reduced FFN defines a function that is identical to the original one on the span of ${p_j}$. This follows directly from the linearity of the second layer and the fact that the value vectors are obtained by querying the original network itself. Therefore, the proposed procedure yields an efficient compression of the parametric memory, with a reduced latent dimensionality $n$, while preserving the dominant non-linear activation behavior of the original FFN.
  
\subsubsection{Algorithm}

The complete compression algorithm is described as follows:

\begin{algorithm}[H]
\caption{Non-Linear SVD for Two-Layer FFN Memory Compression}
\begin{algorithmic}[1] 
\Require Key matrix $K \in \mathbb{R}^{d \times m}$, value matrix $V \in \mathbb{R}^{m \times d}$
\Require target rank $n$; (Optional) dimension retaining threshold $\epsilon$
\Ensure Reduced key matrix $\tilde{K} \in \mathbb{R}^{d \times n}$ and reduced value matrix $\tilde{V} \in \mathbb{R}^{n \times d}$

\State Normalize each column of $K$, extract row norms $\{\alpha_i\}_{i=1}^m$
\State Normalize each column of $V$, extract column norms $\{\beta_i\}_{i=1}^m$
\State Compute composed scalars $s_i = \alpha_i \beta_i$ for all $i$
\State Form weighted key matrix $\hat{K}$ by scaling row $i$ of normalized $K$ by $s_i$

\State Perform EVD on $\hat{K}$: $\hat{K}\hat{K}^\top = U \Sigma U^\top$
\State Select top-$n$ left singular vectors $U_n$

\State Form reduced key matrix $\tilde{K}$ using rows of $U_n$
\State Normalize each row of $\tilde{K}$ (discarding those with norm less than threshold $\epsilon$)

\For{each row vector $\tilde{k_j}$ in $\tilde{K}$}
    \State Feed $\tilde{k_j}$ as input to the original FFN
    \State Collect output vector $\tilde{v_j}$
\EndFor

\State Form reduced value matrix $\tilde{V}$ by stacking $\{\tilde{v_j}\}_{j=1}^n$ as columns \\

\Return $\tilde{K}, \tilde{V}$
\end{algorithmic}
\end{algorithm}

\subsubsection{Expansion-Compression Cycle}

The memory expansion and NL-SVD compression procedures can be composed into a single, coherent expansion-compression cycle that operates over a memorized text span. Given a text span $x_{<T}$, the model processes tokens sequentially. At each token step $t$, the expansion rule is applied independently at each layer. Concretely, for layer $i$, a new key-value pair $(k_m^i, v_m^i)$ is instantiated using the normalized hidden activation $\mathcal{A}_t^i$ and the globally normalized gradient signal $\mathcal{G}_t^i$. This operation monotonically increases the intermediate dimensionality of the FFN.

After processing a contiguous span of $N_{capacity}$ tokens, the accumulated parametric memory may become over-complete. The NL-SVD compression step replaces the oversized FFN with a reduced one whose latent dimensionality is fixed to a target rank $n$ (typically $n=N_{capacity}/2$), while preserving the dominant activation geometry.

From a temporal perspective, the expansion-compression cycle can be applied at flexible granularities. In the extreme case, compression can be invoked after every token, resulting in a fixed-size parametric memory that is continuously refreshed. Alternatively, expansion can be accumulated over longer spans before triggering compression.

\subsubsection{Practical Limitations}
\label{sec:nlsvd_practical}

Despite its theoretical elegance, NL-SVD exhibits notable practical limitations:

\begin{itemize}[leftmargin=15pt]
    \item \textbf{Numerical Precision Requirements}: NL-SVD requires at least single-precision (float32) floating-point arithmetic to ensure numerical stability during the SVD decomposition step. This precludes seamless integration with mixed-precision training schemes (e.g., bfloat16) that are standard practice for efficient LLM training and inference.
    \item \textbf{Computational Overhead}: The SVD operation and related linear algebra routines are not as well-optimized for modern GPU architectures compared to standard gradient-based primitives, resulting in significant computational overhead (12.7$\times$ relative time vs. 1.7$\times$ for BP, as shown in Table~\ref{tab:pg19}).
    \item \textbf{Limited Applicability}: The theoretical guarantees of NL-SVD do not perfectly transfer to Locas-GLU due to its more complex gating mechanism. Since Locas-GLU offers broader compatibility with modern architectures, the practical scope of NL-SVD is limited.
\end{itemize}

Empirical results in Table~\ref{tab:pg19} show that NL-SVD does not demonstrate a clear advantage over BP in terms of final perplexity (17.89 vs. 17.90 at 200K context length), while incurring substantially higher computational cost. Given these considerations, we recommend using BP for memory updates in practice.

\end{document}